\documentclass[twoside,11pt]{article}

\makeatletter\def\input@path{{style/}}\makeatother

\newcommand{\mytitle}{Trade-offs of Local SGD at Scale: An Empirical Study}

\usepackage[preprint]{jmlr2e}

\ShortHeadings{\mytitle}{}
\firstpageno{1}

\title{\mytitle}

\author{\name Jose Javier Gonzalez Ortiz \email josejg@mit.edu\\
\addr MIT CSAIL
\AND
\name Jonathan Frankle \email jfrankle@mit.edu\\
\addr MIT CSAIL
\AND
\name Mike Rabbat \email mikerabbat@fb.com\\
\addr Facebook AI Research
\AND
\name Ari Morcos \email arimorcos@fb.com\\
\addr Facebook AI Research
\AND
\name Nicolas Ballas \email ballasn@fb.com\\
\addr Facebook AI Research}

\editor{Kevin Murphy and Bernhard Sch{\"o}lkopf}

\usepackage{microtype}
\usepackage{graphicx}
\usepackage{caption}
\usepackage{subcaption}
\usepackage{wrapfig}
\usepackage{booktabs}
\usepackage{bm} %
\usepackage{balance}                %
\usepackage{multicol}
\usepackage{setspace}
\usepackage[binary-units=true, per-mode=symbol]{siunitx} %
\usepackage[subtle]{savetrees}    %

\usepackage{etoolbox}
\newtoggle{comments}
\toggletrue{comments} %

\usepackage{xcolor}
\definecolor{ultramarine}{RGB}{0,32,96}
\definecolor{firebrick}{RGB}{178,34,34}
\definecolor{navy}{RGB}{0,0,128}
\definecolor{forestgreen}{RGB}{0,128,0}

\newcommand{\authorcomment}[3]{{\color{#2}[\textbf{#1}:#3]}}
\newcommand{\outline}[1]{\authorcomment{}{ultramarine}{#1}}
\newcommand{\JJ}[1]{\authorcomment{Jose}{firebrick}{#1}}
\newcommand{\mike}[1]{\authorcomment{Mike}{navy}{#1}}
\newcommand{\ari}[1]{\authorcomment{Ari}{blue}{#1}}
\newcommand{\nicolas}[1]{\authorcomment{Nicolas}{forestgreen}{#1}}
\newcommand{\JF}[1]{\authorcomment{Jonathan}{violet}{#1}}

\iftoggle{comments}{}{
    \renewcommand{\outline}[1]{}
    \renewcommand{\JJ}[1]{}
    \renewcommand{\mike}[1]{}
    \renewcommand{\ari}[1]{}
    \renewcommand{\nicolas}[1]{}
    \renewcommand{\JF}[1]{}
}

\usepackage{mathtools} %
\DeclarePairedDelimiter\ceil{\lceil}{\rceil} %
\newcommand{\ts}{\textsuperscript}

\usepackage{algorithm}%
\usepackage[noend]{algpseudocode}
\begin{document}

\maketitle

\begin{abstract}

As datasets and models become increasingly large, distributed training has become a necessary component to allow deep neural networks to train in reasonable amounts of time. 
However, distributed training can have substantial communication overhead that hinders its scalability.
One strategy for reducing this overhead is to perform multiple unsynchronized SGD steps independently on each worker between synchronization steps, a technique known as \emph{local SGD}.
We conduct a comprehensive empirical study of local SGD and related methods on a large scale image classification task.
We find that performing local SGD comes at a price: lower communication costs (and thereby faster training) are accompanied by lower accuracy.
This finding is in contrast from the smaller-scale experiments in prior work, suggesting that local SGD encounters challenges at scale.
We further show that incorporating the slow momentum framework of \citet{wang2019slowmo} consistently improves accuracy without requiring additional communication, hinting at future directions for potentially escaping this trade-off.

\end{abstract}


\section{Introduction} %
\label{sec:introduction}

As datasets and models continue to grow in size, it has become a common practice to train deep neural networks in a distributed manner across multiple hardware \emph{workers} \cite{goyal2017accurate,shallue2018measuring}.
Most deep learning models are currently optimized using some variant of stochastic gradient descent \cite{robbins1951stochastic}, often using a mini-batch approach \cite{bottou2010large,dekel2012optimal}.
In distributed scenarios, the communication overhead necessary to synchronize gradients between workers can quickly dominate the time necessary to compute the model updates, hindering the scalability of this approach.
Moreover, because of the serial nature of neural network training, all the nodes must wait until the synchronization completes, and performance is therefore dependent on the slowest node \cite{dutta2018slow,ferdinand2020anytime}.

These issues have motivated the development of optimization algorithms that reduce the amount of communication between workers.
A simple yet practical example is \emph{local SGD} (closely related to \emph{federated averaging} \citep{mcmahan2017communication}), where, instead of synchronizing the gradients at every iteration, each worker performs multiple SGD steps locally and then averages the model weights across all workers \cite{zhang2016parallel}.
Local SGD has been shown to have good optimization properties from a theoretical standpoint \cite{stich2018local,zhang2016parallel,woodworth2020is}.
However, while local SGD does speed up training, the resulting models are often less accurate compared to a synchronous minibatch SGD baseline \citep{lin2018don}.

\emph{Post-local SGD} is a variant of local SGD introduced by \citet{lin2018don} with the goal of remedying these problems.
Post-local SGD divides training into two phases.
In the first phase, workers perform synchronous minibatch SGD; in the second, they switch to local SGD.
\citeauthor{lin2018don}~claim that this approach generalizes better on large batch training than both local SGD and minibatch SGD while reducing communication for the second phase of training.
The majority of the analysis of local SGD and post-local SGD reported in \citet{lin2018don} is on CIFAR-10 \cite{cifarDsets}.

\begin{figure*}[t]
  \centering
    \includegraphics[width=.97\linewidth]{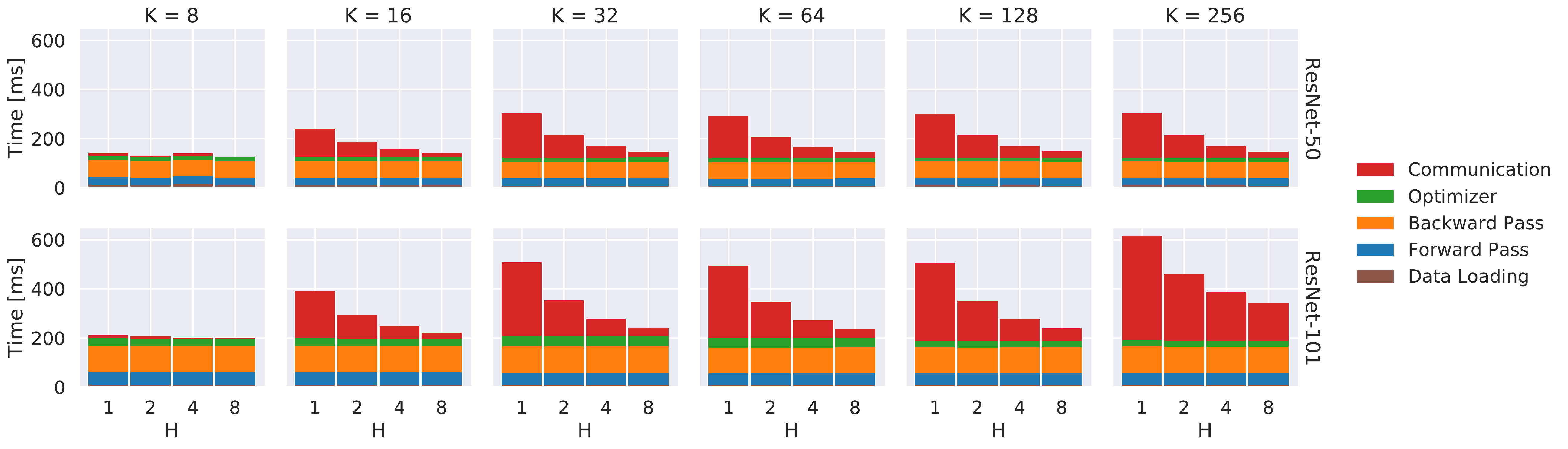}
    \caption{%
        Breakdown of the average wall-clock time per iteration during the training process.
        Results are reported for various numbers of workers ($K$) and numbers of local steps ($H$).
        Every node has 8 workers.
        The main difference is the communication time, which decreases as we reduce the frequency of model averaging.
        For minibatch SGD (which we label as $H=1$) with multiple nodes ($K > 8$), the communication time dominates other parts of training.
       }
    \label{fig:time_split}
\end{figure*}

Motivated by these results, we perform a thorough analysis of local SGD and post-local SGD on the ImageNet-1k \cite{russakovsky2015imagenet} classification task, a \emph{de facto} benchmark for large-scale vision classification problems.
As Figure \ref{fig:time_split} shows, inter-node communication can dominate training time, becoming a bottleneck in the training process.
We complement the analysis of \citet{lin2018don}, studying how the choice of learning rate schedule and the point at which to switch phases affect the generalization accuracy of models trained with post-local SGD.
We find that post-local SGD at ImageNet-scale is a double-edged sword: decreases in communication costs (by increasing the number of local steps) are accompanied by decreases in accuracy.
As a result, practitioners interested in post-local SGD must weigh the trade-offs between training speedup and reduction in the accuracy of the final model.
Looking ahead, our analysis of the interaction between post-local SGD, learning rates, and momentum points toward potential opportunities to escape these trade-offs.

\textbf{Contributions}. Our main contributions are as follows:
\begin{enumerate}
    \vspace{-2mm}
    \itemsep.5pt%
    \item We perform a comprehensive empirical study on ImageNet that identifies previously unreported scalability limitations of local and post-local SGD. Our analysis highlights how, when compared to the fully synchronous baseline, local and post-local SGD suffer from non-trivial accuracy drops as workers or local steps increase.
    \item Our analysis is the first to identify that post-local SGD performance heavily relies on the choice of hyperparameters, including learning rate schedule and switching point.
    \item We show that using slow momentum \cite{wang2019slowmo} together with post-local SGD achieves a better  quality-performance trade-off. %
    \item We show that switching to local SGD has a regularization effect on optimization that is only beneficial in the short term, suggesting it is always better to make the switch later in training.

\end{enumerate}

\section{Background and Related Work} %
\label{sec:related_work}

\textbf{Minibatch SGD.}
Neural networks are typically trained with \emph{minibatch SGD}.
In minibatch SGD, the dataset $\mathcal{D} = \{(x_i, y_i)\}_{i \in [N]}$ is divided into non-overlapping subsets
of size $B$ known as \emph{minibatches}.
Gradient descent is performed sequentially on these minibatches, passing through the entire dataset over the course of an \emph{epoch}.
The dataset is typically randomly shuffled before each epoch, meaning the minibatch composition and order are different on each pass through the dataset.
See Algorithm \ref{alg:minibatch-sgd} below for full details.
In practice, networks are often trained in a \emph{distributed} data-parallel fashion across $K$ workers \cite{li2020pytorch}.
Each worker has a separate copy of the weights and computes gradients using a disjoint subset of the data.
The entire dataset is reshuffled and split among workers at the beginning of every epoch.
After the backward pass, gradients are averaged across workers before updating the model weights.

\textbf{Local SGD.}
In local SGD, described in Algorithm~\ref{alg:local-sgd}, the workers update the local copies of their weights, and every $H > 1$ iterations they \emph{synchronize} the weights across workers by averaging the weights stored on each worker.
Papers on local SGD typically credit \citet{mcdonald2009efficient}, \citet{zinkevich2010parallelized} and \citet{mcdonald2010distributed} with pioneering local SGD in pre-deep learning settings; these works train models to completion and average the final parameters.
\citet{zhang2016parallel} explore local SGD with periodic averaging of models throughout training; they prove that it converges in convex settings (in fact, faster than minibatch SGD) in the face of gradient variance and show that it can optimize a LeNet-5 network on MNIST.
\citet{povey2014parallel} and \citet{su2015experiments} use local natural gradient descent with periodic synchronization to optimize deep models.
\citet{kamp2018efficient} synchronize updates among neighboring workers using a gossiping protocol.
\citet{zhou2017convergence}, \citet{stich2018local}, \citet{wang2018cooperative}, \citet{yu2019parallel}, and \citet{woodworth2020is} prove convergence properties of local SGD under various technical assumptions. \citet{zhou2017convergence} experiment on CIFAR-10 and show that accuracy drops off if synchronization occurs too infrequently.
\citet{wang2018adaptive} start with infrequent averaging and then increase the communication over the course of training.
Periodic model averaging also takes place in \emph{federated learning}, which focuses on performing SGD in settings where data is distributed across many devices and a model can only be updated on the device where the data resides. %
\citet{mcmahan2017communication}, who initiated this research literature, perform parallel SGD between devices with synchronization on every step, although they explore synchronizing after many iterations.
Local SGD is common in federated learning systems for performance reasons \citep{smith2017federated}. %

\textbf{Post-local SGD.}
Post-local SGD \citep{lin2018don} is a hybrid in which training occurs synchronously for the first part of training and switches to local SGD later in training.
Post-local SGD involves performing synchronous SGD for the first $T$ steps of training and local SGD thereafter.
The hyperparameters may differ between the synchronous and local phases.
For example, the local phase may use a different learning rate or batch size.
\citeauthor{lin2018don}~suggest post-local SGD as a replacement for standard, synchronous, large-batch training schemes that ``significantly improves the generalization performance.''

Other approaches for reducing communication costs include using gradient compression to reduce the size of updates \citep{alistarh2017qsgd,wen2017terngrad,bernstein2018signsgd,karimireddy2019error,vogels2019powersgd} and modifying the communication patterns between nodes \cite{lian2017can,lian2018asynchronous,assran2019stochastic}, often by introducing approximation or asynchrony.

\begin{algorithm}[t]
\small
\caption{Minibatch SGD with learning rate schedule $\gamma(t)$, batch size $B$, initial weights $w_0$, and loss $\ell$.}
\label{alg:minibatch-sgd}
\begin{spacing}{1.1}
\begin{algorithmic}[1]
\State $S \leftarrow \ceil{\frac{N}{B}}$ (iterations per epoch)
\For{each epoch $e$}
\State Shuffle dataset $\mathcal{D} = \{x_i, y_i\}_{i \in N}$
\For{each iteration $m$ in $1$ to $S$}
\State $t \leftarrow eS + m$ (the current step of training)
\State $o \leftarrow (m-1)B$
\State $w_t \leftarrow w_{t-1} - \gamma(t) \frac{1}{B}\sum_{i=1}^B \nabla \ell(f(x_{o+i}; w_t), y_{o+i})$
\EndFor
\EndFor
\end{algorithmic}
\end{spacing}
\end{algorithm}

\begin{algorithm}[t]
\small
\caption{Local SGD with learning rate schedule $\gamma(t)$, batch size $B$, initial weights $w_0$, loss $\ell$, and $R$ workers.}
\label{alg:local-sgd}
\begin{spacing}{1.1}
\begin{algorithmic}[1]

\State $S \leftarrow\ceil{\frac{N}{BK}}$ (iterations per epoch)
\For{each epoch $e$}
\State Shuffle dataset $\mathcal{D} = \{x_i, y_i\}_{i \in N}$
\For{each iteration $m$ in $1$ to $S$, at worker $k$ (in parallel)}
\State $t \leftarrow eS + m$ (the current step of training)
\State $o \leftarrow (m-1)B + (k-1)SB$
\State $w_t^{(k)} \leftarrow w^{(k)}_{t-1} - \gamma(t) \frac{1}{B}\sum_{i=1}^B \nabla \ell(f(x_{o+i}; w_t), y_{o+i})$
\If{$t \bmod H = 0$}
\State $w^{(k)}_{t} \leftarrow \frac{1}{K} \sum_{k=1}^K w^{(k)}_{t}$ (synchronize)
\EndIf
\EndFor
\EndFor
\end{algorithmic}
\end{spacing}
\end{algorithm}

\section{Experimental Setup} %
\label{sec:experimental_results}

Our aim in this work is to experimentally investigate the performance of post-local SGD and the related methods on a large-scale distributed training workload.
Here, we describe the experimental setup for the experiments.
Further implementation details are listed in Appendix \ref{platform_details}

\textbf{Dataset}.
We focus on the computer vision task of image classification on the ImageNet dataset \citep{deng2009imagenet,russakovsky2015imagenet}.
This is a 1000-way classification task with $\sim$1.28 million training images. Performance is evaluated by top-1 error on 50,000 held out validation images.
For the training examples, we follow the standard channel normalization and data augmentation scheme as in \citet{resnet2} and use the default training and validation splits from \citet{russakovsky2015imagenet}.

\textbf{Models}.
We use the ResNet-50 and ResNet-101 architectures \citep{resnet}.
We initialize the weights as in \citet{he2015delving} and incorporate the modifications
described in \citet{goyal2017accurate} for initializing the fully connected layer and the
last Batch Normalization layer of the residual blocks.
We use L2 weight decay with $\lambda = 10^{-4}$.

\textbf{Training}.
Following recommendations from \citet{goyal2017accurate}, we scale the learning rate linearly with the global batch size according to $\eta = K B \,0.1/256 $, where $K$ is the number of workers and $B$ is the per-worker batch size, which is set to $32$.
In addition, we apply a learning rate warm-up strategy that starts from $0.1$ and linearly increases the
learning rate every iteration during the first five epochs of training so that the target learning rate $\eta$ is reached at the end of the fifth epoch.
All experiments are trained with Nesterov momentum \citep{nesterov2013introductory} with $\beta = 0.9$.
We also apply momentum correction as described in \citet{goyal2017accurate} to stabilize training whenever the learning rate value is modified.
Unless mentioned otherwise, we use a step-wise learning rate schedule that decays the learning rate by a factor of 10 at the end of the 30\ts{th}, 60\ts{th} and 80\ts{th} epochs, as in \citet{resnet}.

\textbf{Local SGD}.
For local SGD and post-local SGD, we average the model weights of all the workers before computing the validation performance at the end of every epoch. This model average is performed outside of the training process.
We consistently found that averaging the models before computing validation was beneficial to generalization performance.
For post-local SGD, before the switch to local SGD is performed, we average the gradients and not the model weights. %
During the post-local SGD regime the per-worker momentum buffers are kept local and not synchronized, this is a common approach \cite{lian2017can,assran2019stochastic,wang2019slowmo}.

\section{The Benefits of (Post-)Local SGD} %
\label{sec:scalability_of_local_sgd}

While local SGD undoubtedly reduces communication and thus overall runtime, the specific speedup varies from task to task.
In this section we illustrate the scalability challenges of minibatch SGD in distributed settings and show that local SGD and post-local SGD achieve substantially better performance by reducing communication.
To reduce the effect of run-to-run variation, we execute ten independent runs (seeds) and report the median performance values.
We consider local SGD where the model is averaged every $H$ steps from the beginning of training and post-local SGD with a switching point at the first learning rate decay (epoch 30) as suggested in \citet{lin2018don}.

\textbf{Local SGD reduces communication overhead}.
The main performance cost of scaling minibatch SGD to a distributed setting is the communication overhead to synchronize gradients before updating model weights at every iteration.
If this overhead is comparable to the other parts of training, the training process will be significantly slowed down, leading to reduced performance.
We show that, for the classification task considered on a \SI{10}{Gb\per s} Ethernet interconnect connection, the performance benefits of local SGD are substantial.

Figure~\ref{fig:time_split} presents a breakdown of the running times of different parts of the training loop for a variety of settings. Communication time refers to performing an all-reduce to average the weights across all workers.
As expected, all except for the communication time stay approximately constant for different numbers of workers.
For a single node, i.e., 8 GPUs, the communication time is quite low, indicating that intra-node communication is negligible with respect to other parts of training.
However, for multi-node minibatch SGD ($H=1$ and $K>8$ in the figure), the communication time dominates the other steps as the number of workers $K$ increases.
On the other hand, for local SGD ($H>1$ in the figure) as we increase the number of local steps $H$ the communication overhead is amortized over several iterations, leading to a reduced iteration time.

\begin{figure*}
\begin{minipage}[t]{0.49\linewidth}
  \centering
    \includegraphics[width=\linewidth]{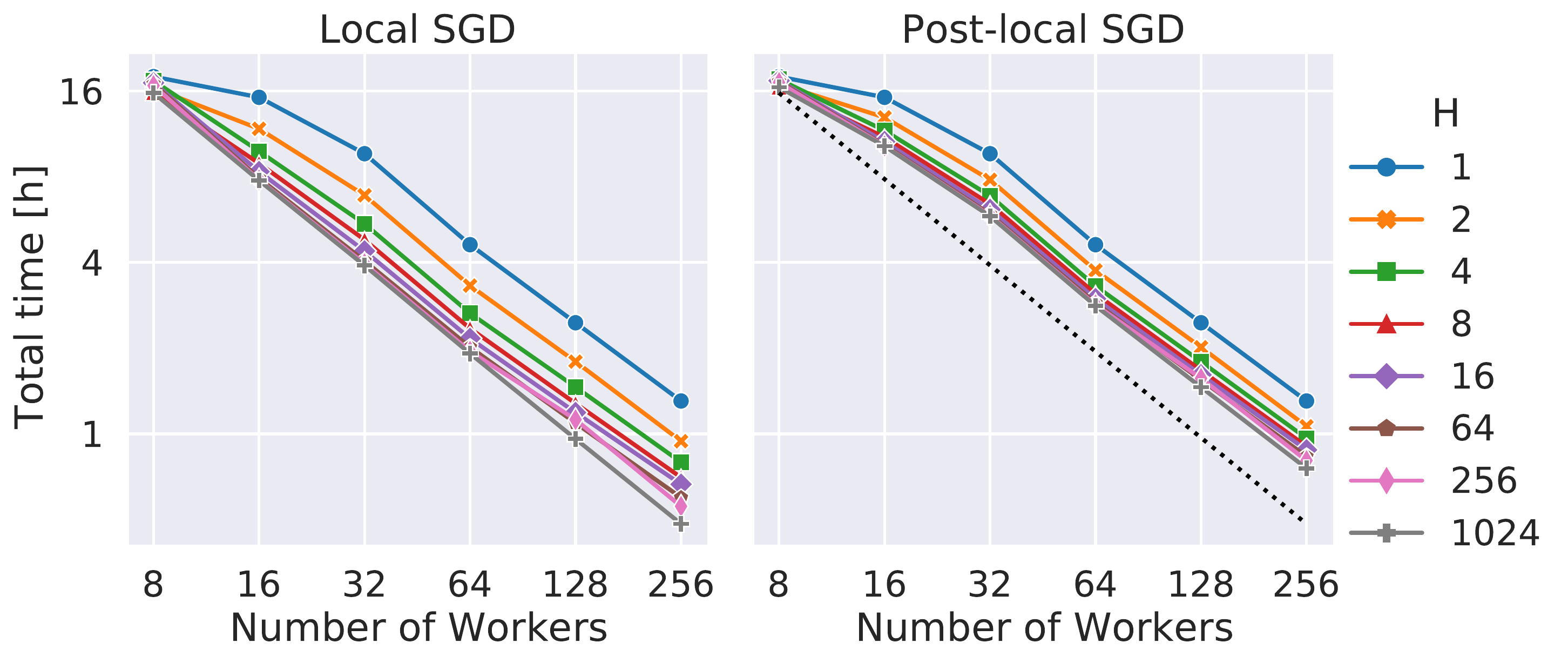}
     \caption{%
    Total time in hours as a function of the number of workers for local SGD and post-local SGD.
    We report median time over 10 separate runs.
    }
    \label{fig:time_per_epoch}
\end{minipage}
\hfill
\begin{minipage}[t]{0.49\linewidth}
  \centering
    \includegraphics[width=\linewidth]{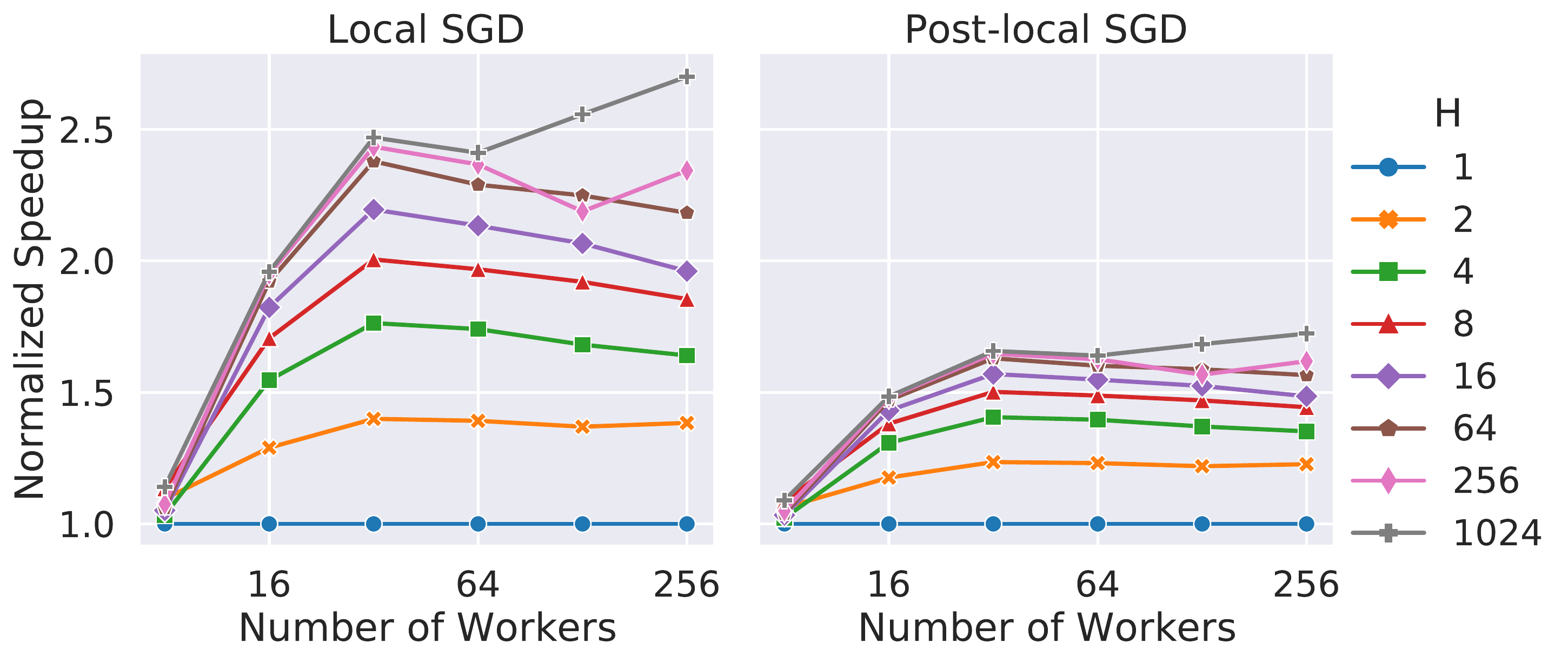}
    \caption{%
    Normalized Speedup of local and post-local SGD for a given world size.
    I.e., the ratio between the time per epoch for local SGD and the corresponding minibatch SGD.%
    }
    \label{fig:normalized_speedup}
\end{minipage}
\end{figure*}

\textbf{Local SGD scales to large distributed settings}. %
While we see that communication overhead is a large fraction of the time per iteration in Figure~\ref{fig:time_split}, we also observe that the communication time is not proportional to the number of workers.
Consequently, despite the introduced communication overhead when going from single-node to multi-node training, we should observe good scalability trends as we increase the overall number of workers.
This is depicted in Figure~\ref{fig:time_per_epoch}, where the total time, in hours, to complete 90 epochs of training a ResNet-50 is presented for local and post-local SGD for different numbers of workers and local steps.
The approximately linear trends indicate that we can successfully make use of additional nodes to achieve reduced training times without running into other performance bottlenecks.
In general, we observe that in both minibatch SGD (indicated as $H = 1$ in Figure \ref{fig:time_per_epoch}) and local SGD, doubling the number of workers halves the time taken to run an epoch.
Whereas for minibatch SGD going from a single node to multiple nodes (i.e., 8 to 16 workers) has an effect on running time, local SGD shows better scalability asymptotically approaching linear speedup. %

\textbf{Reducing communication is effective but has diminishing returns}. %
While we observed overall training time reductions, it is still unclear what specific speedup reducing communication is accounting for.
To measure the contribution of local SGD, we normalize times when $H > 1$ by the time for $H=1$ with the same number of workers, resulting in the normalized speedup shown in Figure~\ref{fig:normalized_speedup}.
This normalized speedup corresponds to the speedup obtained by using local SGD with $H > 1$ relative to using $H=1$ (i.e., synchronizing every iteration, equivalent to minibatch SGD) with the same number of workers.
With more than one node ($K > 8$), local SGD provides a substantial speedup. For instance, 32 or 64 workers communicating every 4 or 8 local steps leads to relative speedups close to 1.75 and 2 respectively.
However, as we keep increasing the number of steps between synchronizations, the speedup increases with diminishing returns since the amount of synchronizations decays exponentially.

\begin{figure*}[t]

\begin{minipage}[t]{0.49\linewidth}
  \centering
    \includegraphics[width=\linewidth]{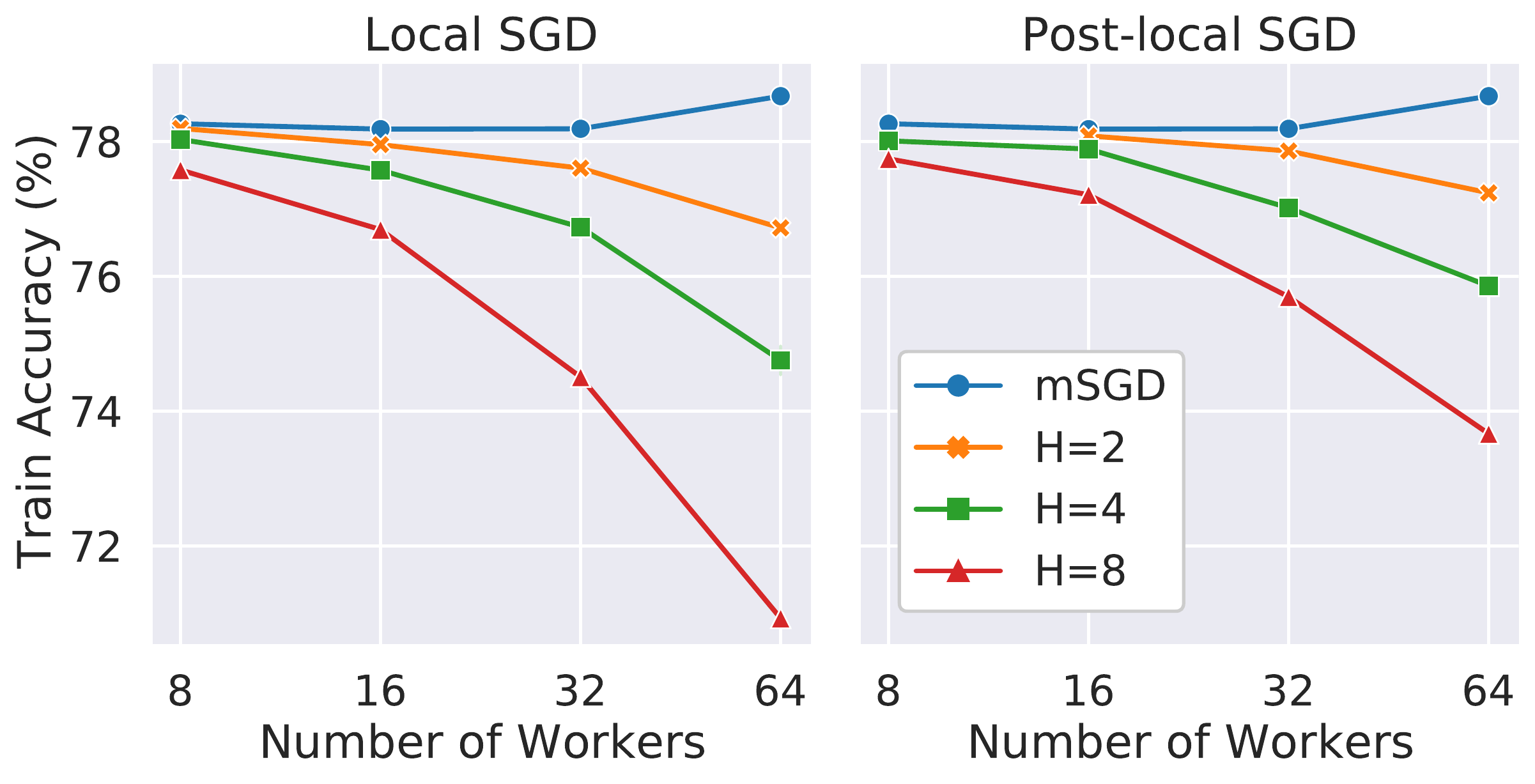}
    \caption{%
    Training Accuracy as a function of the number of workers for local SGD (left) and post-local SGD (right).
    Separate curves refer to experiments with the same number of local steps $H$ between model averages.
    mSGD corresponds to the minibatch SGD baseline.
    }
    \label{fig:T0-30_vs_K_train}
\end{minipage}
\hfill
\begin{minipage}[t]{0.49\linewidth}
  \centering
    \includegraphics[width=\linewidth]{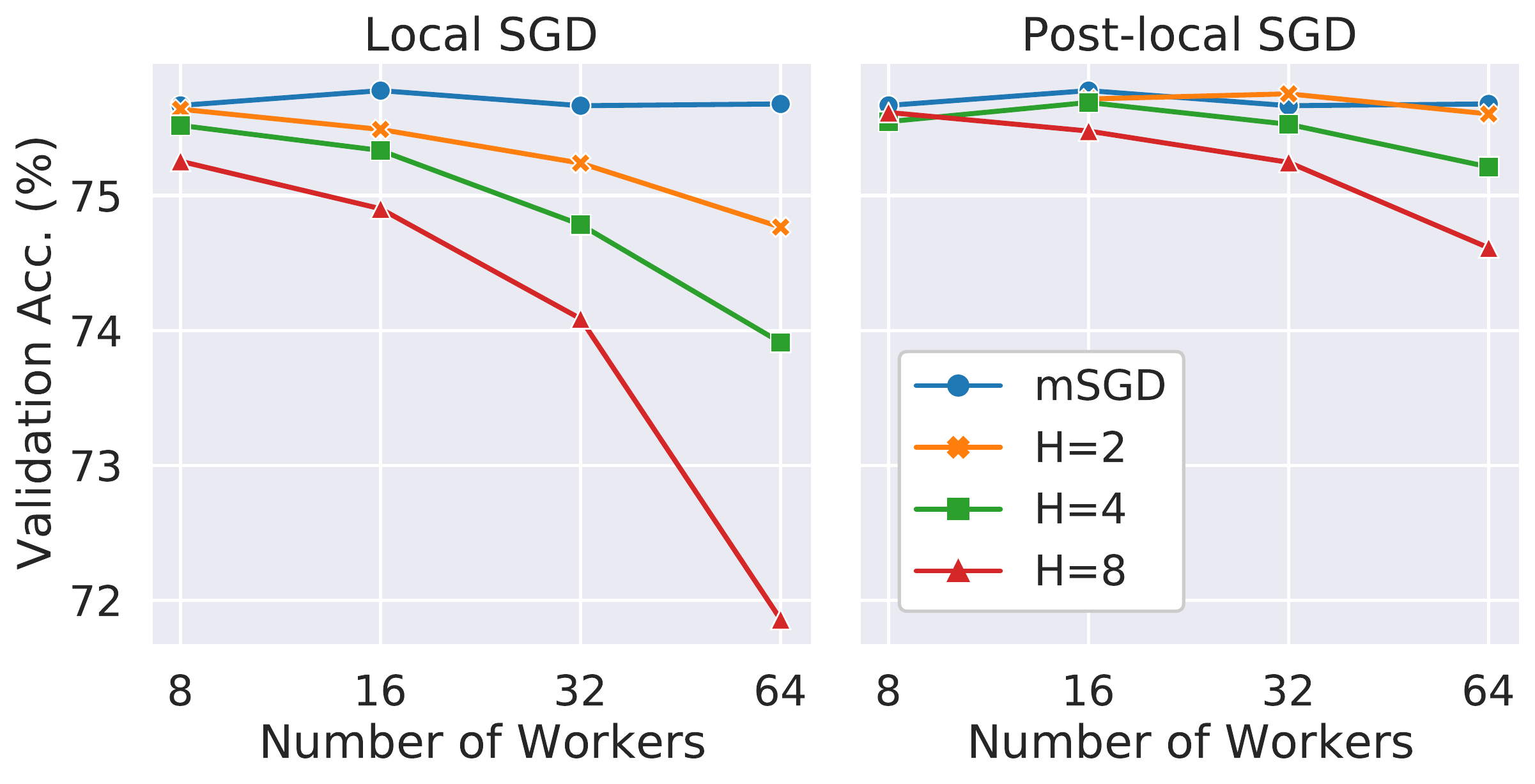}
    \caption{%
    Validation Accuracy as a function of the number of workers for local SGD (left) and post-local SGD (right).
    Separate curves refer to experiments with the same number of local steps $H$ between model averages.
    mSGD corresponds to the minibatch SGD baseline.
    }
    \label{fig:T0-30_vs_K}
\end{minipage}
\end{figure*}

\section{Trade-offs of (Post-)Local SGD} %
\label{sec:challenges_of_scaling_}
\subsection{Trade-offs of local model updates}

The measured speedup benefits of reducing communication during the training process suggest that (post-)local SGD presents an opportunity to substantially reduce training times.
As such, we now analyze the effect on generalization of these techniques to understand whether this performance comes at a cost.
We explore the scalability of both local SGD and post-local SGD (switching at epoch 30 unless otherwise noted) as we increase the total number of workers (i.e., GPUs) and as we increase the number of local steps between model averaging.
We further study varying the amount of communication of post-local SGD by changing the switching point between the phases. %
We show that on ImageNet-1k, local SGD and post-local SGD struggle to scale in any of these axes without causing a drop in validation performance.

\textbf{Increasing the number of workers}.
Our first analysis focuses on the trade-off between increasing the number of workers and the effects on generalization as measured by the decrease in validation accuracy.
We find that increasing workers has a negative effect on accuracy for both local and post-local SGD, consistently leading to worse training and validation results.
Figures \ref{fig:T0-30_vs_K_train} and \ref{fig:T0-30_vs_K} show training and validation accuracy as the number of workers increases.
We include results for various numbers of local steps.
Neither local SGD nor post-local SGD maintain accuracy as the number of workers increases.
Although post-local SGD experiences smaller absolute drops in accuracy compared to local SGD, this comes at the expense of more communication and consequently more runtime.
The loss in accuracy seems to be a direct result of the optimization mechanics of local SGD, since the minibatch baseline
does not experience this trade-off as the number of workers increases.

\textbf{Reducing synchronization frequency}. As we saw from the earlier timing results, reducing the frequency of synchronizations is the main mechanism that local SGD has for reducing the communication overhead.
Here, we investigate the effect of increasing the number of local steps between model averaging synchronizations, identifying a very similar trade-off to increasing the number of workers.
Figures  \ref{fig:T0-30_vs_K_train} and \ref{fig:T0-30_vs_K} show training and validation accuracy for varying numbers of local steps $H$ for local and post-local SGD.
For both methods, as the number of local steps grows larger, training and validation accuracy monotonically decrease.

\textbf{Towards a unified trade-off}.
We showed that increasing the number of workers $K$ or local steps $H$ led to lower accuracy.
Interestingly, neither of these factors was clearly dominant and the absolute drop in training and validation accuracy is similar as we increase either of them.
We show that this phenomenon can be better viewed in terms of a different statistic: total number of local model updates between synchronizations, i.e., the product $K \times H$.

Both of the factors $K$ and $H$ affect the optimization mechanics.
As we increase the number of workers, $K$, the global batch size linearly increases because every worker has a fixed local batch size.
    Consequently, since the number of epochs is fixed, the total number of model updates decreases by the same factor.
    In local SGD models are averaged after $H$ local steps. While model averaging usually leads to better performance, it depends on how much the local models have diverged since the last synchronization.

\begin{wrapfigure}{L}{0.52\textwidth}
  \centering
    \includegraphics[width=\linewidth]{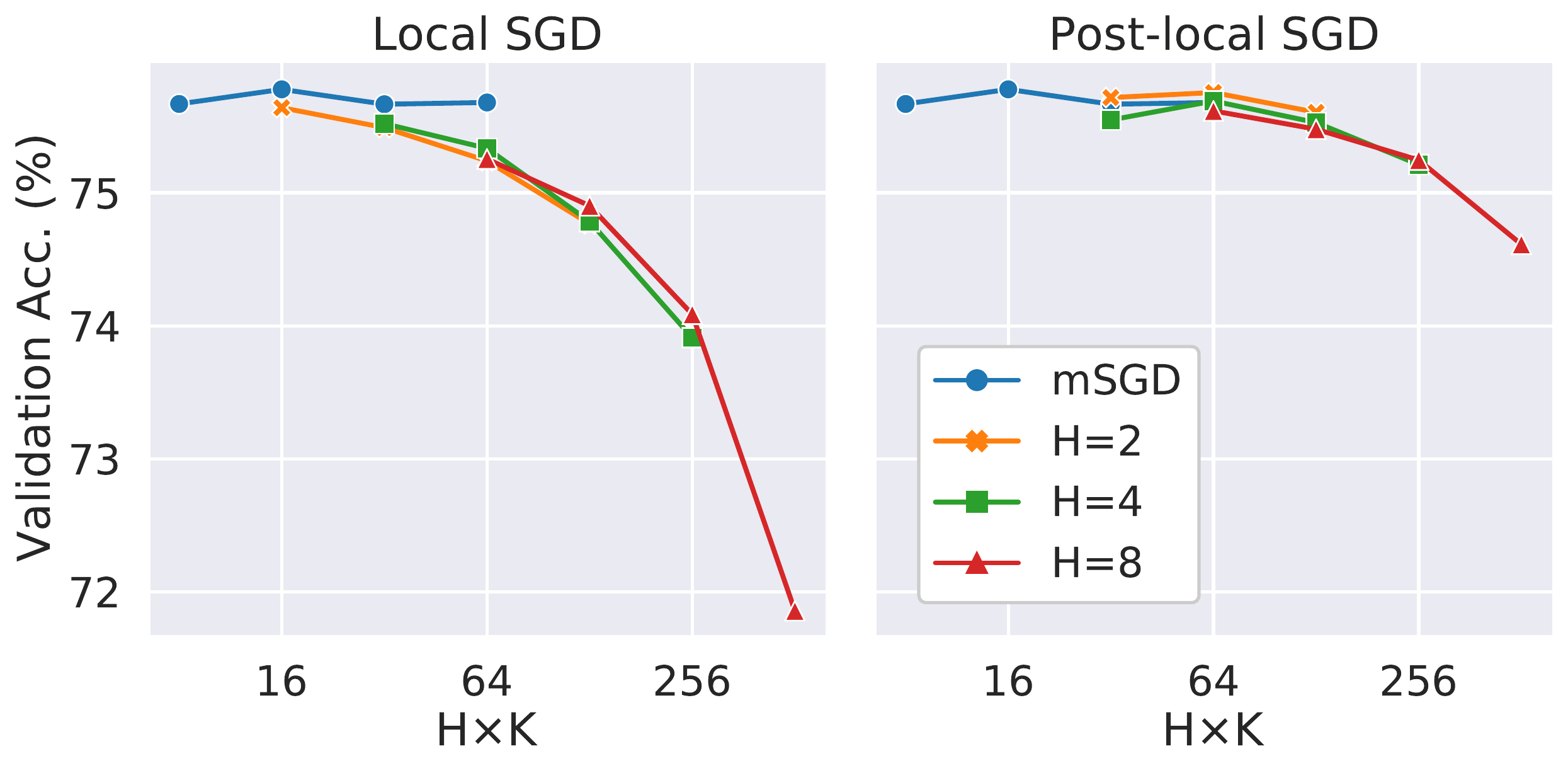}
    \caption{%
     Validation Accuracy as a function of the number of local model updates ($H\times K$) for local SGD (left) and post-local SGD (right). mSGD values correspond to the minibatch SGD baseline.
    }
    \label{fig:T0-30_vs_HK}
\end{wrapfigure}

Figure \ref{fig:T0-30_vs_HK} shows validation accuracy as a function of the product $H \times K$. We found the same pattern holds for training accuracy (see Figure \ref{fig:T0-30_vs_HK_train} in the Appendix).
When viewed this way, we can observe that the trade-off curves for different numbers of workers and frequencies line up for both training and validation.
This suggests that the main factor influencing final model quality is the total number of local model updates between synchronizations.
Consequently, to maintain model accuracy, an increase in the number of local steps should be matched with a proportional decrease in the number of workers and \emph{vice versa}.

\subsection{The effect of hyperparameters}

\textbf{The post-local switching point poses a trade-off}.
Post-local SGD divides training into two phases. In the first phase, workers perform synchronous minibatch SGD; in the second, they switch to local SGD.
However, the iteration at which to switch is another hyperparameter that affects the amount of communication; here, we study whether it also affects final accuracy.
We find that the switching point presents a trade-off between training time and final accuracy, with later switching points improving accuracy at the cost of additional communication.

Up to this point, all the reported experiments switched to local SGD at epoch 30 out of 90. This is when the first learning rate decay happens with a step-wise schedule with decays at epochs 30, 60 and 80
(as recommended by \citet{lin2018don}).
However, switching earlier can reduce the amount of communication.
Conversely, previous results showed consistent improvements from local SGD to post-local SGD and, since local SGD can be thought of as a post-local SGD with a switching point at epoch 0, we also explore whether we can recover the lost model accuracy by switching at a later point.

\begin{figure*}[t]
\begin{minipage}[t]{0.49\linewidth}
  \centering
    \includegraphics[width=\linewidth]{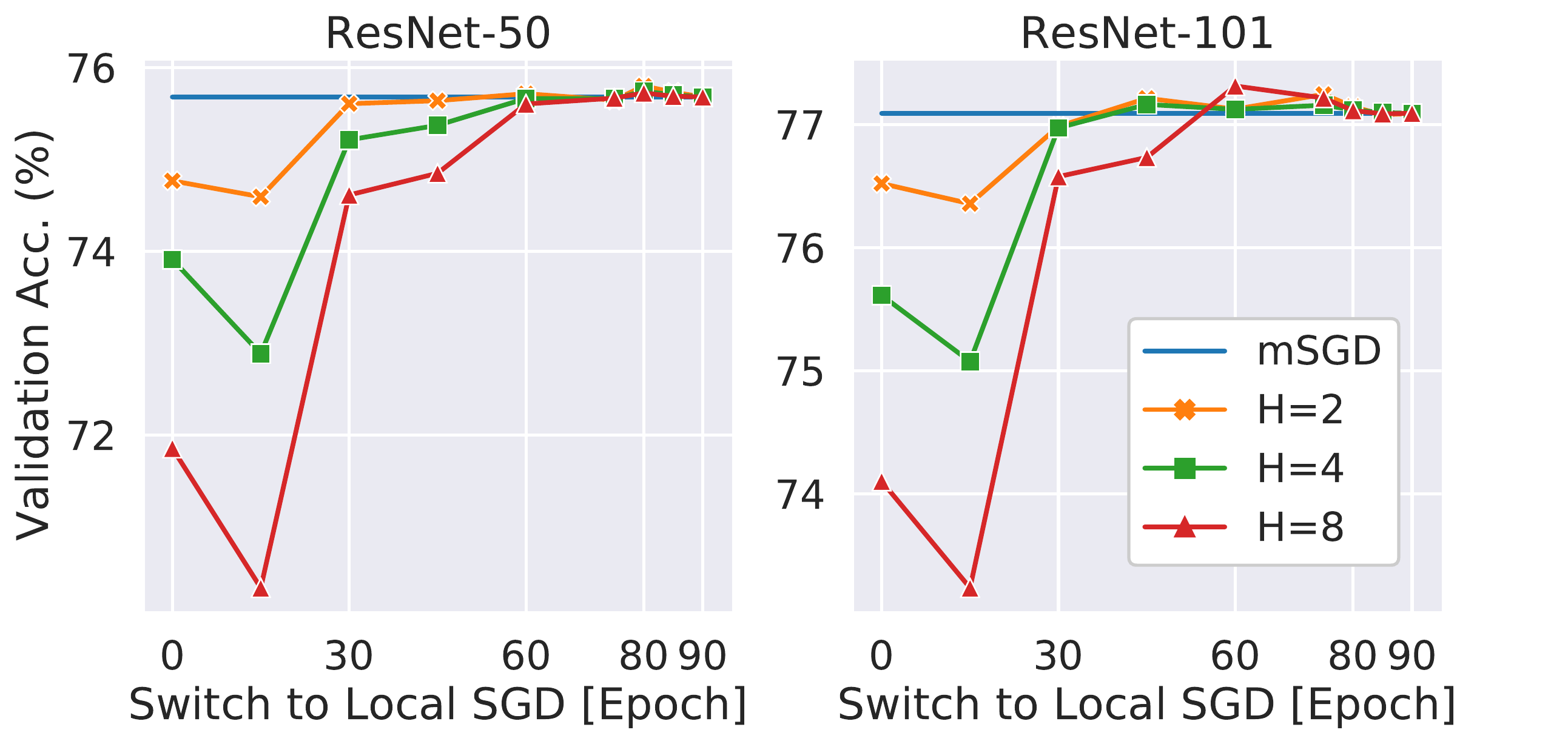}
    \caption{%
    Validation accuracy for post-local SGD with step-wise learning rate schedule as a function of the epoch when the switch to local SGD is done.
    Results are for $K=64$ workers and various numbers of local steps $H$. mSGD is the baseline minibatch SGD method, synchronizing at every update.
    }
    \label{fig:TVA_vs_T_multi}
\end{minipage}
\hfill
\begin{minipage}[t]{0.49\linewidth}
  \centering
    \includegraphics[width=\linewidth]{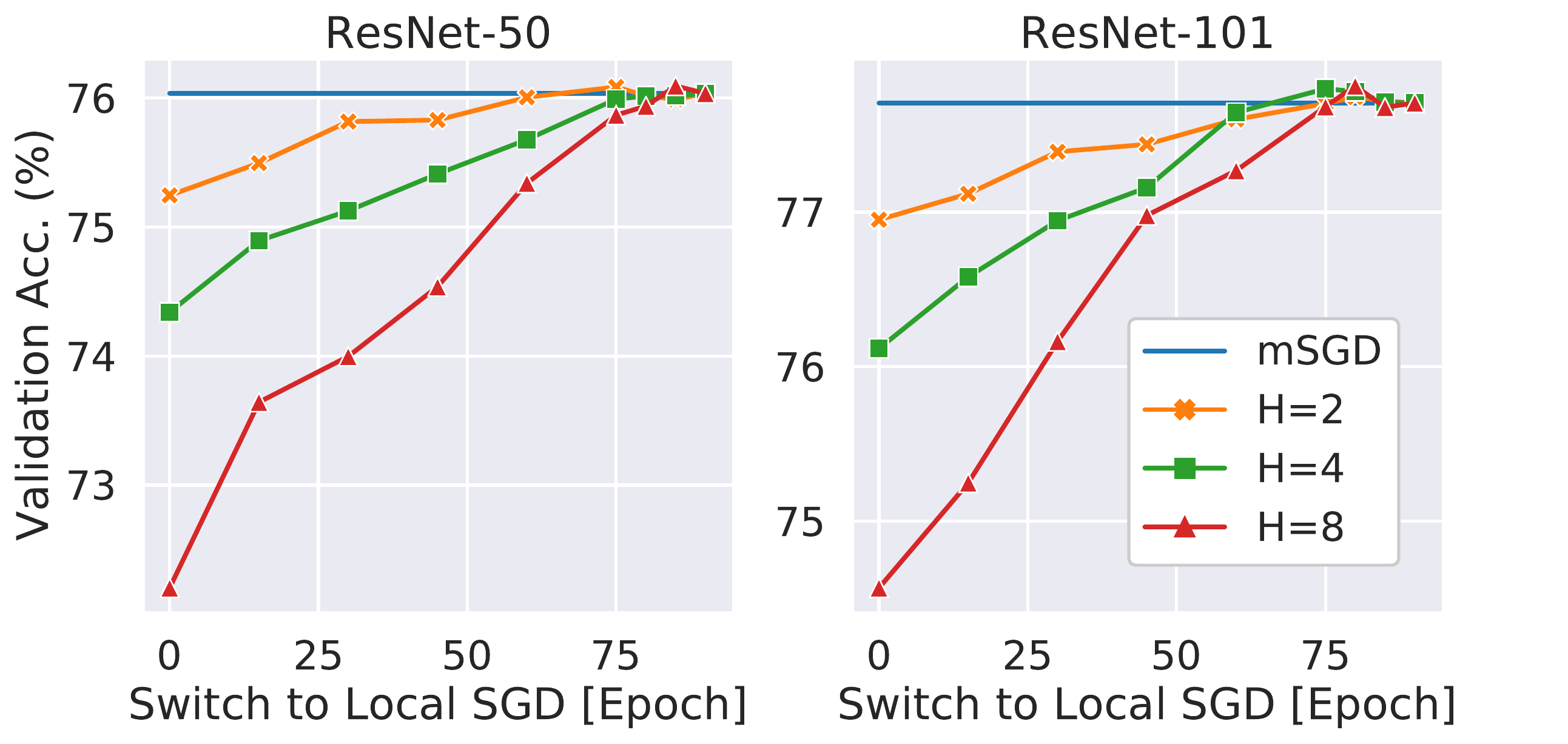}
    \caption{%
    Validation accuracy for post-local SGD with half cosine learning rate schedule as a function of the epoch when the switch to local SGD is done.
    Results are for $K=64$ workers and various numbers of local steps $H$. mSGD is the baseline minibatch SGD method, synchronizing at every update.
    }
    \label{fig:TVA_vs_T_cosine}
\end{minipage}

\end{figure*}

To study this phenomena, we select switching points at epochs where the learning rate decreases (30, 60 and 80) and intermediate epochs (15, 45, 75, and 85).
Figure \ref{fig:TVA_vs_T_multi} depicts these results for both ResNet-50 and ResNet-101 for various numbers of local steps.
Both networks present similar patterns for the chosen values of switching points.
Notably, the learning rate decay points seem to have a significant positive effect into the final performance for this task.

\textbf{Post-local SGD performance heavily depends on learning rate schedule}.
One unexplored aspect of post-local SGD is the interaction between the switching time heuristic and the choice of learning rate schedule.
Here, we investigate how the choice of learning rate schedule affects the generalization results of post-local SGD. %
Recently, learning rate schedules other than step-wise, such as cosine annealing, have gained popularity due to the competitive results they provide without having to carefully specify the learning rate decay points \citep{he2019bag,radosavovic2019network}.
As an alternative to the step-wise schedule, we consider a half-period cosine schedule which sets the learning rate to $\eta_{t}= \eta_0 (1+\cos(\pi \cdot t/(T_\text{max}))/2$, where $\eta_0$ is the initial learning rate, $t$ is the current epoch, and $T_\text{max}$ is the total number of epochs.
As with the step-wise schedule, we modify the cosine schedule to have a linear warm-up phase updated every iteration over the first five epochs of training.
The learning rate is updated at the end of every epoch except for the warm-up phase, when it is updated at every iteration.

Figure \ref{fig:TVA_vs_T_cosine} presents a trade-off analysis of how the switching point affects the final training accuracy of the model with a half cosine learning rate schedule.
Unlike before,  we see a more monotonic trade-off between the final validation accuracy and when the switch to local SGD is performed.
When compared to Figure \ref{fig:TVA_vs_T_multi}, the results in Figure \ref{fig:TVA_vs_T_cosine} indicate that the scale of the learning rate when the switch to local SGD is performed is crucial to ensure that the drop in validation accuracy is minimal.

\begin{figure}[t]

\end{figure}

\begin{figure*}[t]
\begin{minipage}[t]{0.46\linewidth}
   \centering
    \includegraphics[width=.95\linewidth]{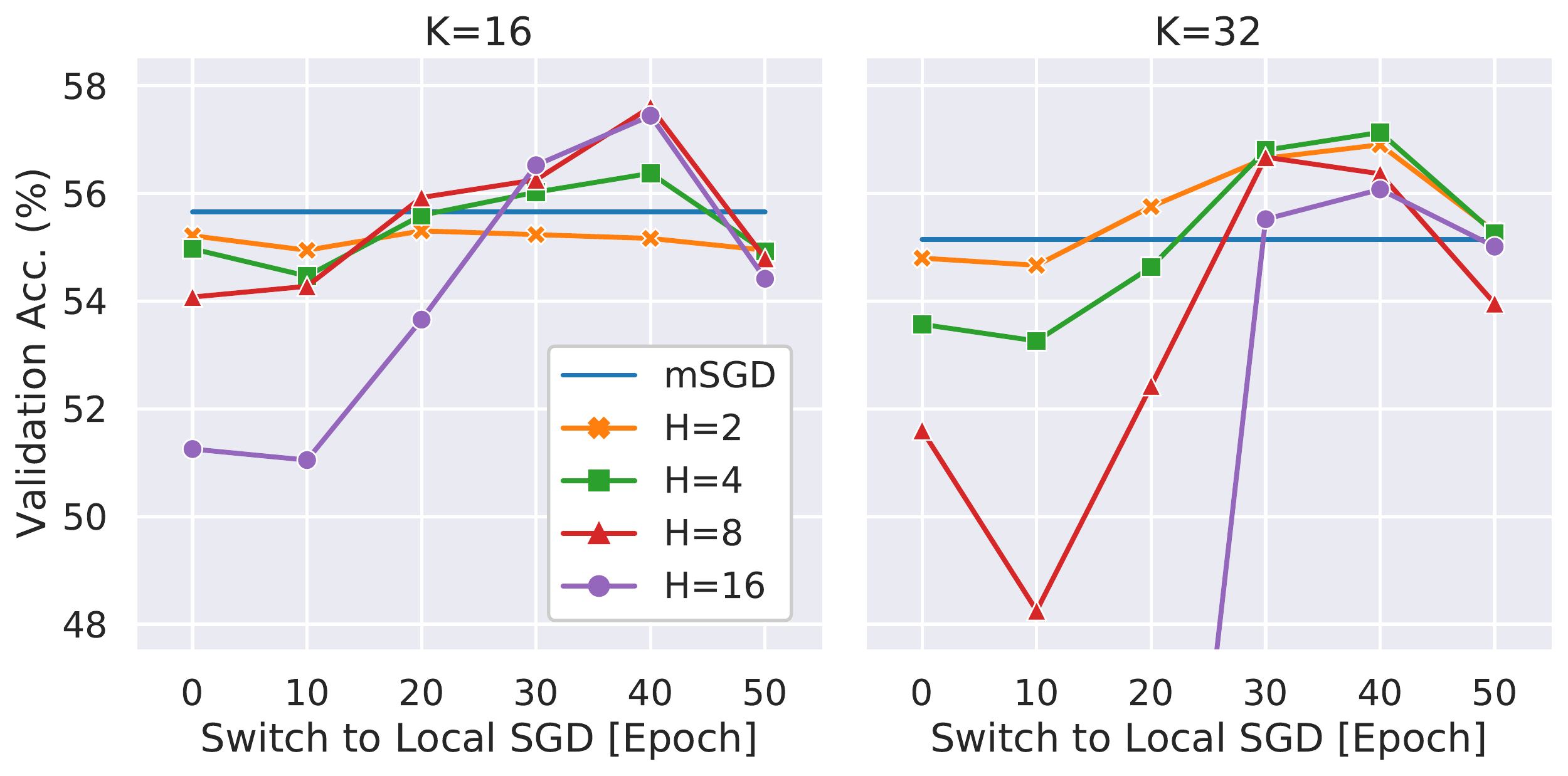}
    \caption{%
    Validation accuracy for post-local SGD on TinyImageNet as a function of the epoch when switching to local SGD. Learning rate is decayed at epochs 30 and 50.
    }
    \label{fig:postlocal-tiny}
\end{minipage}
\hfill
\begin{minipage}[t]{0.52\linewidth}
  \centering
    \includegraphics[width=\linewidth]{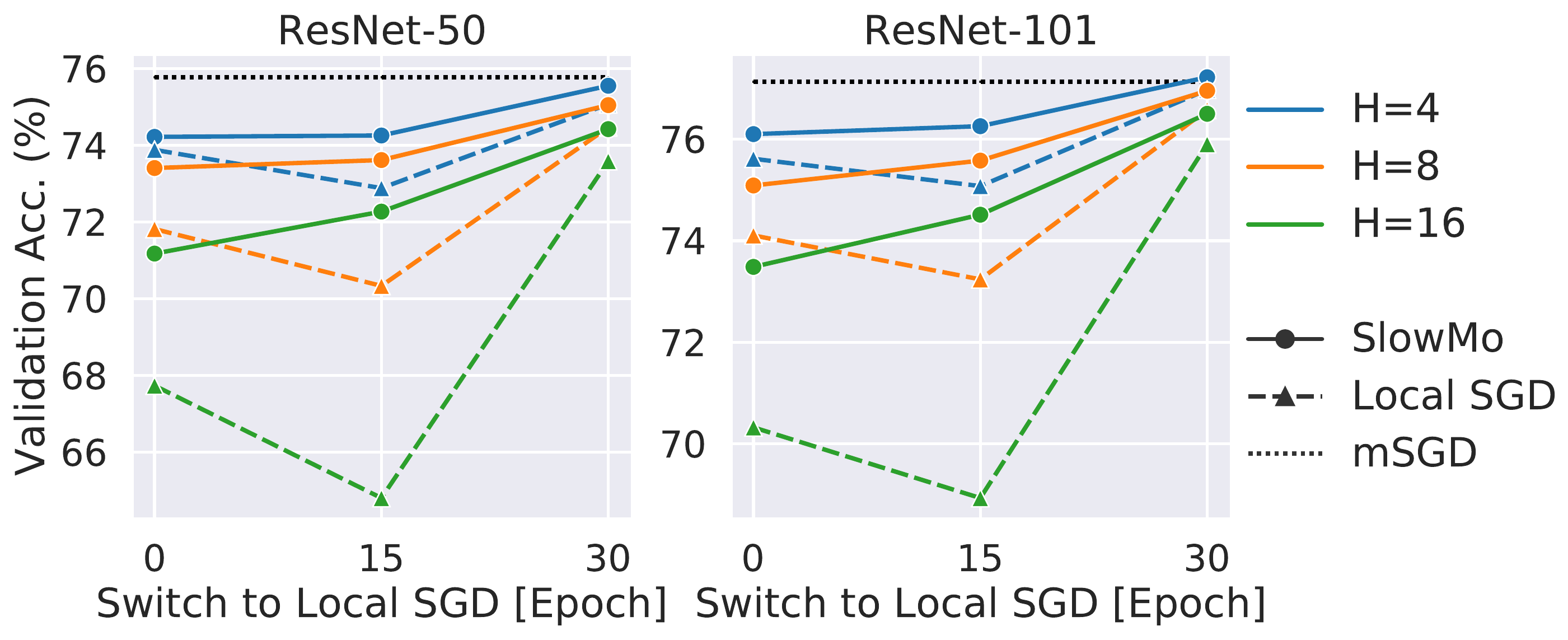}
    \caption{%
    Validation accuracy for various local SGD settings with and without the slow momentum update.
    SlowMo yields better performance than local or post-local SGD.
    }
    \label{fig:postlocal-slowmo}
\end{minipage}

\end{figure*}

\subsection{Post-local SGD behavior depends on the task scale} %
\label{sub:post_local_sgd_b}

Post-local SGD was proposed as a way to remedy the generalization gap present in large batch training \cite{chen2016scalable,keskar2016large,hoffer2017train,shallue2018measuring}.
However, our observations do not align with the generalization improvements identified by \citet{lin2018don} for ResNet-20 on CIFAR-10 for increasing values of $K$ and $H$ even when exploring different switching points.
We found similar patterns for a downscaled version of the previously presented experiments.
We performed analogous experiments on TinyImageNet, a dataset which comprises a subset of ImageNet at a lower resolution, using a ResNet-50 (with slight modifications to work on the smaller dataset).
Figure \ref{fig:postlocal-tiny} shows that post-local SGD (when performed after the first learning rate decay point) does produce accuracy gains with respect to the synchronous minibatch baseline.
This contrasts with our ImageNet experiments, where little to no accuracy gains are achieved when switching to local SGD during training.
We have thus identified an example where distributed optimization behaves differently for tasks of different scales.

We note that our ImageNet results are not directly comparable to the results of \citet{lin2018don} on ImageNet since we do not use the LARS optimizer \citep{you2017large} for the main body of experiments; we use SGD with Nesterov momentum instead.
To ensure that choice of optimizer is not a confounding factor in this analysis, we performed analogous experiments using LARS and the results are included in Figure \ref{fig:lars} in the Appendix.
Our analysis indicates that post-local SGD + LARS suffers from a similar trade-off.
When switching to local SGD in the early phase of training, final model accuracy drops with respect to the minibatch baseline.
In contrast, we do see that LARS trade-off is more linear and provides some accuracy improvements when the switch occurs later in training.

\section{Shifting the Trade-off with Slow Momentum}

We have identified several trade-offs of local and post-local SGD in terms of training time and model accuracy.
We now explore modifying the optimization process as a way to mitigate the drop in performance that local-SGD and post-local SGD experience, arriving at better trade-off.
To that end we make use of the slow momentum (SlowMo) method~\cite{wang2019slowmo}, which builds on top of the \textrm{BMUF} method \citep{chen2016scalable}  by generalizing local SGD in a way similar to how SGD with momentum generalizes SGD.
In particular, the slow momentum algorithm works by adding an additional momentum update after model averaging steps.
SlowMo consistently provided improvements when incorporated into algorithms with reduced communication, including local SGD and other decentralized algorithms.
SlowMo does not require additional communication since the model averages provide enough information to compute the slow gradients and perform the update.

We evaluate SlowMo on top of local SGD and post-local SGD with switching points at 15 or 30 epochs.
Figure \ref{fig:postlocal-slowmo} shows that adding the slow momentum updates leads to improved validation accuracy in all cases.
Incorporating SlowMo at the switching point at epoch 15 leads to  noticeably better accuracy, outperforming the local SGD case and inverting the behavior previously seen.
Although SlowMo is not able to fully bridge the accuracy gap compared to synchronous minibatch SGD, it does mitigate the drop by reliably boosting performance for different settings.

\section{The Effect of Local SGD on Generalization} %
\label{sec:local_sgd_effect_on_generalization}

Up to this point, our analysis has focused on the accuracy achieved by post-local SGD.
However, the best validation accuracy of the model might not be representative of how local SGD and post-local SGD affect the optimization of the model over the course of training.
Here, we study how the loss and the errors evolve over time for different switching points revealing that switching to local-SGD has a regularization effect regardless of when it is performed.
We show that switching to local SGD has a regularization effect on optimization that is only beneficial in the short term, suggesting it is always better to make the switch later in training.

\begin{figure*}
  \centering
    \includegraphics[width=\linewidth]{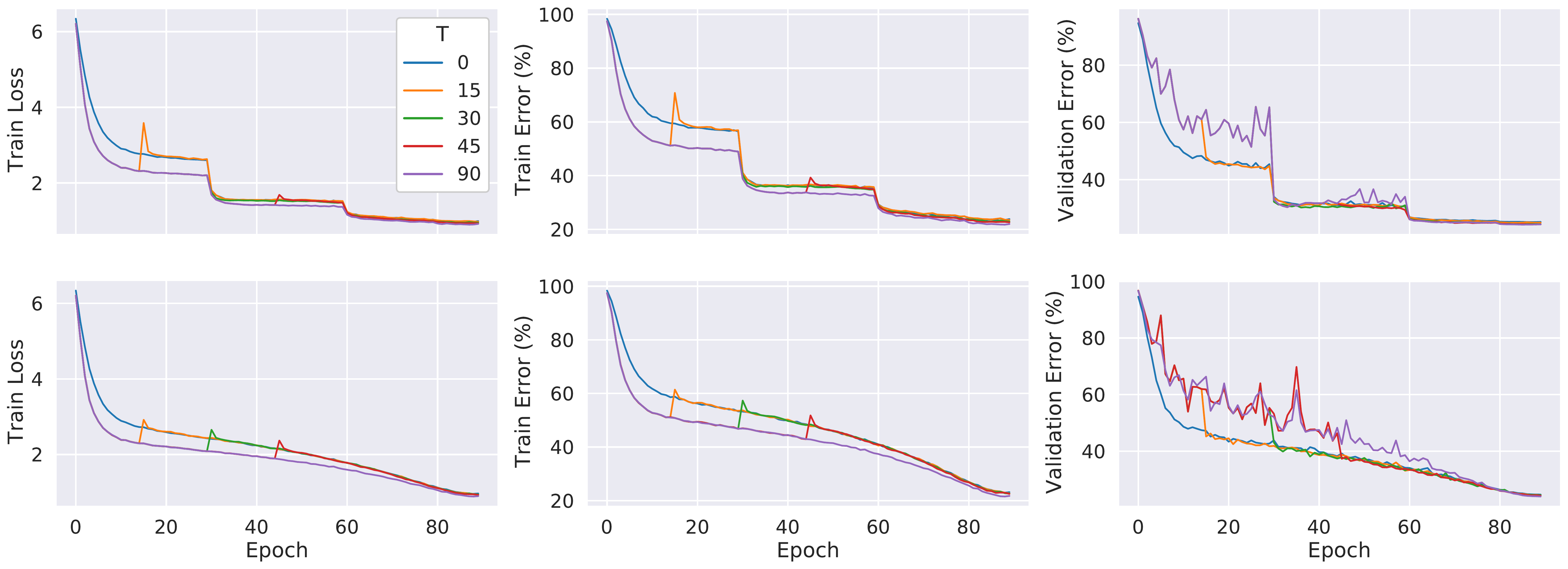}
    \caption{%
    Training Loss (first column), Training Error (second column) and Validation Error (third column) for
    post-local SGD experiments with different switching points $T$ and number of local steps $H=4$.
    Top row corresponds to step-wise learning rate schedule with
    decays at 30, 60 and 80 epochs.
    Bottom row corresponds to a half cosine schedule.
    For a given number of local steps, regardless of when the switch to local SGD is performed, the loss and error curves follow the same general trajectory.
    We found this pattern consistently across different models, numbers of workers and numbers of local steps.
    }
    \label{fig:loss_tracks}
\end{figure*}

In Figure  \ref{fig:loss_tracks} we plot the training loss, top-1 training error and top-1 validation error for post-local SGD for various switching points. $T=0$ corresponds to local-SGD and $T=90$ corresponds to the minibatch baseline.
We observe a regularization effect when the switch to local SGD is performed for the post-local SGD method: the training error is higher than the minibatch baseline while the validation error is lower.
Results are for 64 workers and 4 local steps between model averages but we found these patterns consistently for different numbers of workers and number of local steps.
More interestingly, after the switching point, the curves' various post-local SGD models closely follow the same trajectory for both training and validation metrics.
This pattern holds for both learning rate schedules although it is more noticeable for the half cosine learning rate schedule because of the higher absolute change.

If we now consider these results along with the earlier analysis that looked at the final validation accuracy, we can conclude that, despite the switch to local-SGD improving generalization in the short term, the long term effects are detrimental to final model accuracy.
We believe that this interplay has not been studied thoroughly before and that it
appears to be a critical component when making use of strategies that
switch from a fully synchronous regime to local training, like
post-local SGD.

\section{Limitations}

A limitation of this work is that it only studies the ImageNet-1k classification task.
Nevertheless, with over 1.28 million images and 1000 classes, ImageNet-1k is considered a \emph{de facto} benchmark for natural image vision classification tasks. Our experiments study a wide variety of settings and account for more than 47,000 GPU hours of compute.
Moreover, existing analysis on CIFAR-10 along with our experiments using TinyImageNet seem to indicate that the behavior of post-local SGD depends on the task scale. Since methods such as local SGD are used in a distributed setting mostly required by large networks and datasets, we deem important that a benchmark of this size is used.
Lastly, while we do not propose new algorithmic approaches, our analysis identifies trade-offs of local and post-local SGD not reported before which impact its applicability by researchers and practitioners alike. For instance, the dependence between final model accuracy and factors like learning rate or switching point is critical to the viability of post-local SGD.

\section{Conclusion}
\label{sec:conclusion}

In this paper we present a thorough analysis of local and post-local SGD at scale by evaluating them on the ImageNet-1k classification task.
We identify several scalability limitations that local and post-local SGD experience and analyze the trade-offs they present in terms of final model accuracy.
We find that the number of workers and the number of local steps for local SGD present accuracy trade-offs, and we unify these hyperparameters into a trade-off expressed in terms of number of local model updates between synchronizations.
We characterize the behavior of post-local SGD for different switching points along the training process, revealing a similar  trade-off between communication and model accuracy.
Furthermore, we identify that the choice of learning rate schedule has a large impact for the resulting accuracy of post-local SGD.
We further show that incorporating the slow momentum framework of \citet{wang2019slowmo} consistently improves accuracy without requiring additional communication. %
Lastly, our analysis reveals that switching to local SGD has a regularization effect on optimization that is beneficial in the short term but detrimental to final model accuracy.

\vskip 0.2in
\bibliography{%
references/architectures,%
references/datasets,%
references/deeplearning,%
references/distributed,%
references/decentralized,%
references/gradient-compression,%
references/largebatch,%
references/local-sgd,%
references/optimizers,%
references/packages,%
references/sgd,%
references/vision-cls%
}

\clearpage
\appendix
\clearpage
\section{Experimental Setup - Additional Details} %

\subsection{LARS} %
\label{sub:lars}

We make use of the LARS optimizer as implemented in NVIDIA APEX library with clipping enabled and a trust coefficient of 0.02 and $\epsilon = 10^{-8}$for all the reported experiments.
Despite LARS being proposed with a polynomial learning rate schedule, we use the same step-wise learning rate schedule as with the other experiments to reduce the chance of confounding factors.
We found the polynomial schedule to perform slightly better ($< 0.5\%$) than the step-wise but both are able to achieve competitive top 1 validation accuracy results.
All other hyperparameters are identical to the setup described for SGD with momentum.

\subsection{TinyImageNet} %
\label{ssub:tinyimagenet}

As discussed in the main body of the paper, we perform experiments on TinyImageNet, a spatially downscaled subset of ImageNet.
Here we include details of our experimental setup for results involving TinyImageNet.
Unlike with other datasets, to our knowledge there is no rigorous analysis like \citet{goyal2017accurate} for TinyImageNet that contains validated recommendations of hyperparameters.
Therefore, we performed a series of hyperparameter searches and arrived at the following choices after cross-validating several settings.

We use a slightly modified version of ResNet-50 and ResNet-101 where the first maxpool layer has been discarded (as in \citet{leavitt2020selectivity}) to account for the difference in spatial size of the images (224x224 in ImageNet-1k vs 64x64 in TinyImageNet).
Similarly, we halve the size of the learnable filters to account to the reduction in dataset complexity.
Models are trained for 60 epochs with learning rate decays at epochs 30 and 50.
We use L2 weight decay with $\lambda = 10^{-3}$. We found weight decay to be one of the most influential hyperparameters for this task.
Apart from these changes, the rest of the training setup uses the same choices as with the ImageNet experiments: learning rate warmup, momentum correction, learning rate scaling and model initialization.

\subsection{SlowMo}

SlowMo introduces two additional hyperparameters: $\alpha$, the ``slow'' learning rate, and $\beta$, the ``slow'' momentum coefficient.
For computational reasons, we do not sweep the $\alpha$ and $\beta$ hyperparameters of SlowMo and set them to recommended defaults of $\alpha=1$ and $\beta=0.5$.

\subsection{Platform Details}
\label{platform_details}

All methods are implemented using PyTorch 1.6 \citep{pytorch-neurips} and we use the ResNet implementations from torchvision 0.7 with CUDA 10.1 and the NCCL communication library.
Our experiments run on nodes with eight NVIDIA V100 GPUs each. The nodes communicate over \SI{10}{Gb\per s} Ethernet links.
Throughout, a worker refers to a process in a node that makes use of one GPU exclusively.
Thus, the number of workers is equivalent to the number of GPUs and, in our setup, is equivalent to eight times the number of nodes.

\newpage
\section{Additional Experiments} %
\label{sec:additional_results}

\begin{figure}[ht]
  \centering
    \includegraphics[width=0.7\linewidth]{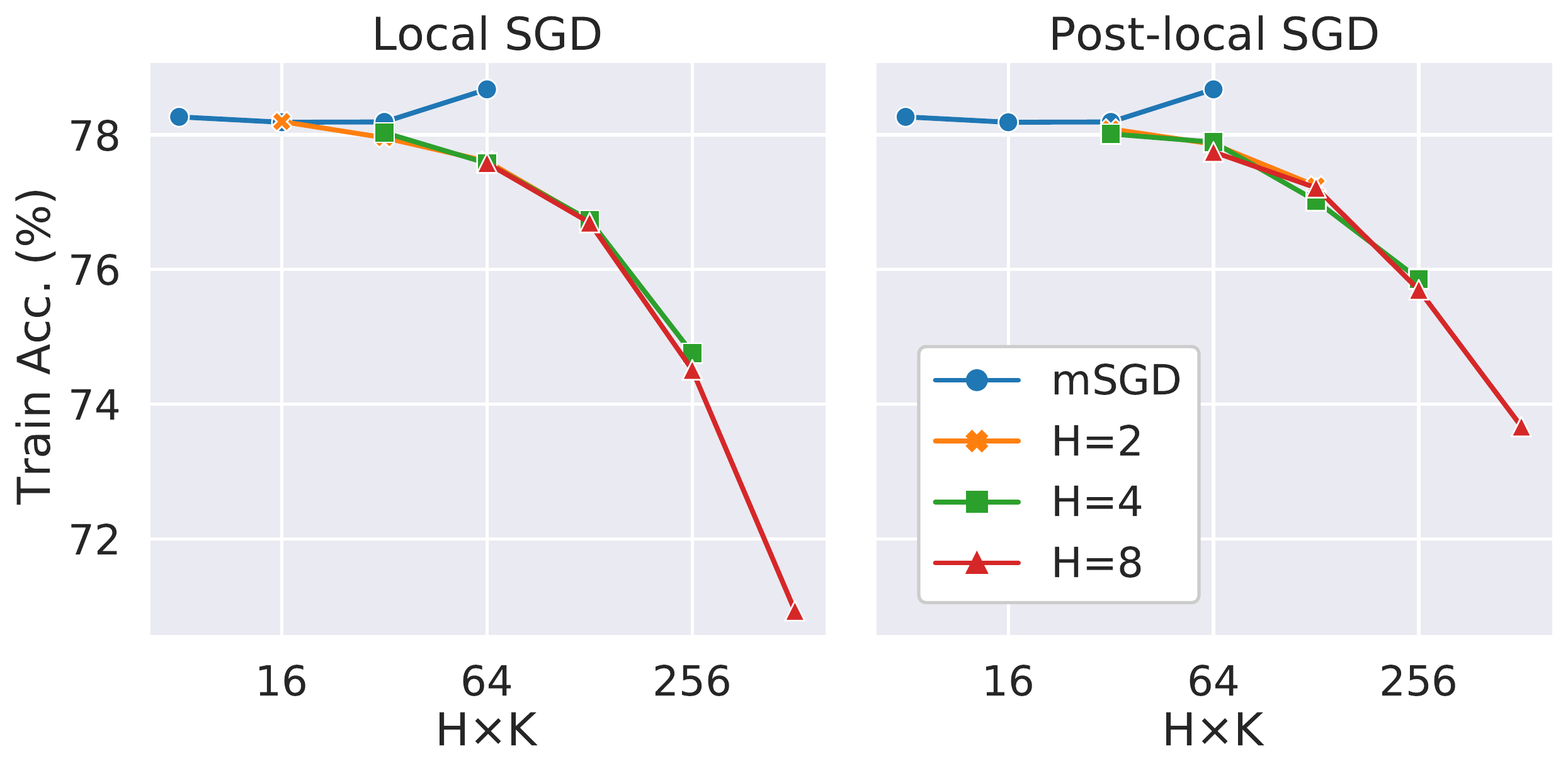}
    \caption{%
     Training Accuracy as a function of the number of local model updates ($H\cdot K$)  for local SGD (left) and post-local SGD (right). mSGD values correspond to the minibatch SGD baseline.
    }
    \label{fig:T0-30_vs_HK_train}
\end{figure}

\begin{figure}[ht]
  \centering
    \includegraphics[width=0.7\linewidth]{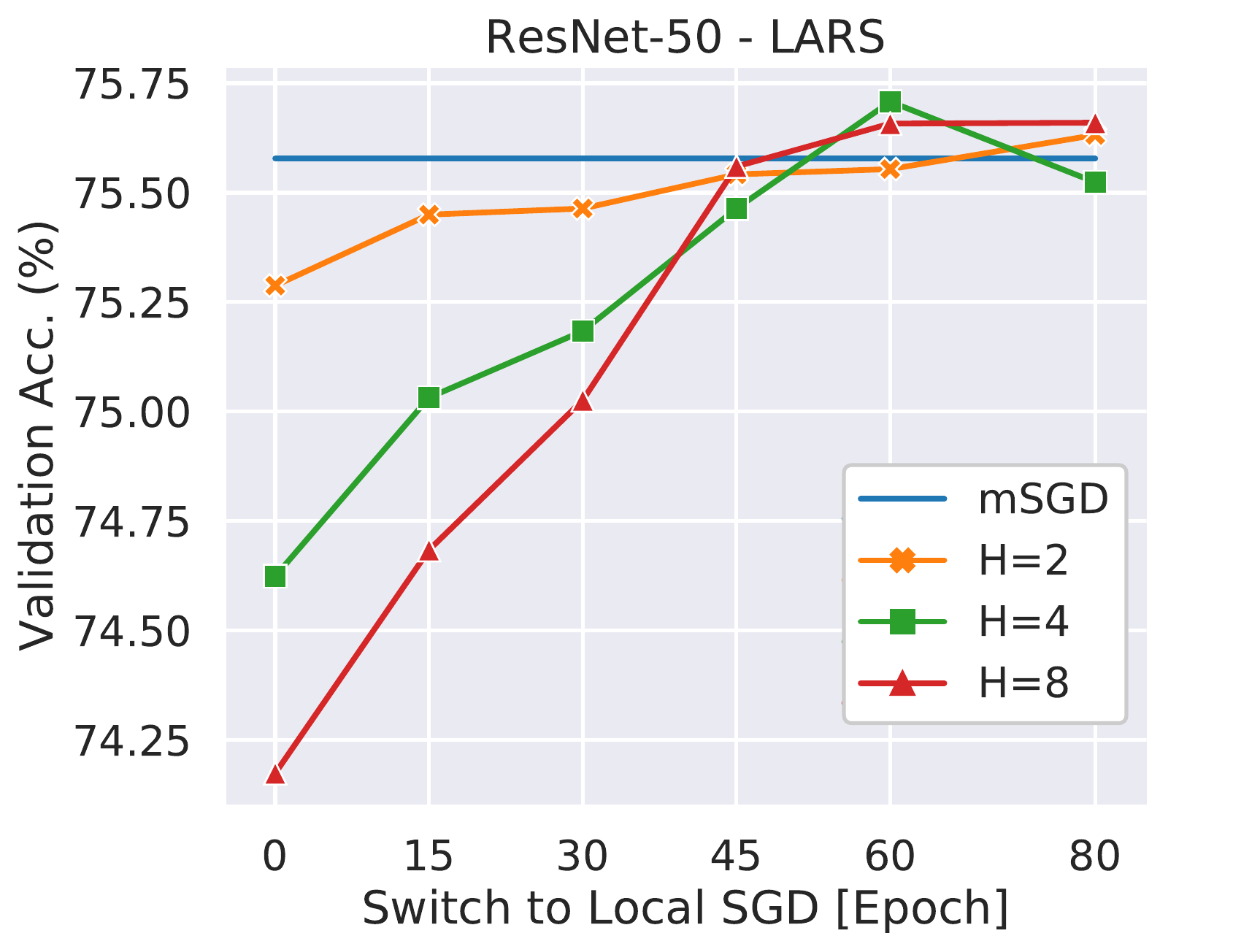}
    \caption{%
        Validation accuracy for post-local SGD with the LARS optimizer as a function of the epoch when the switch to local SGD is done. Results are for $K=32$ workers and step-wise learning rate schedule. LARS presents a similar trade-off to SGD, with switching points later into training providing better accuracy. LARS seems to have a more linear trade-off without a direct impact.
    }
    \label{fig:lars}
\end{figure}

\begin{figure}[ht]
  \centering
    \includegraphics[width=\linewidth]{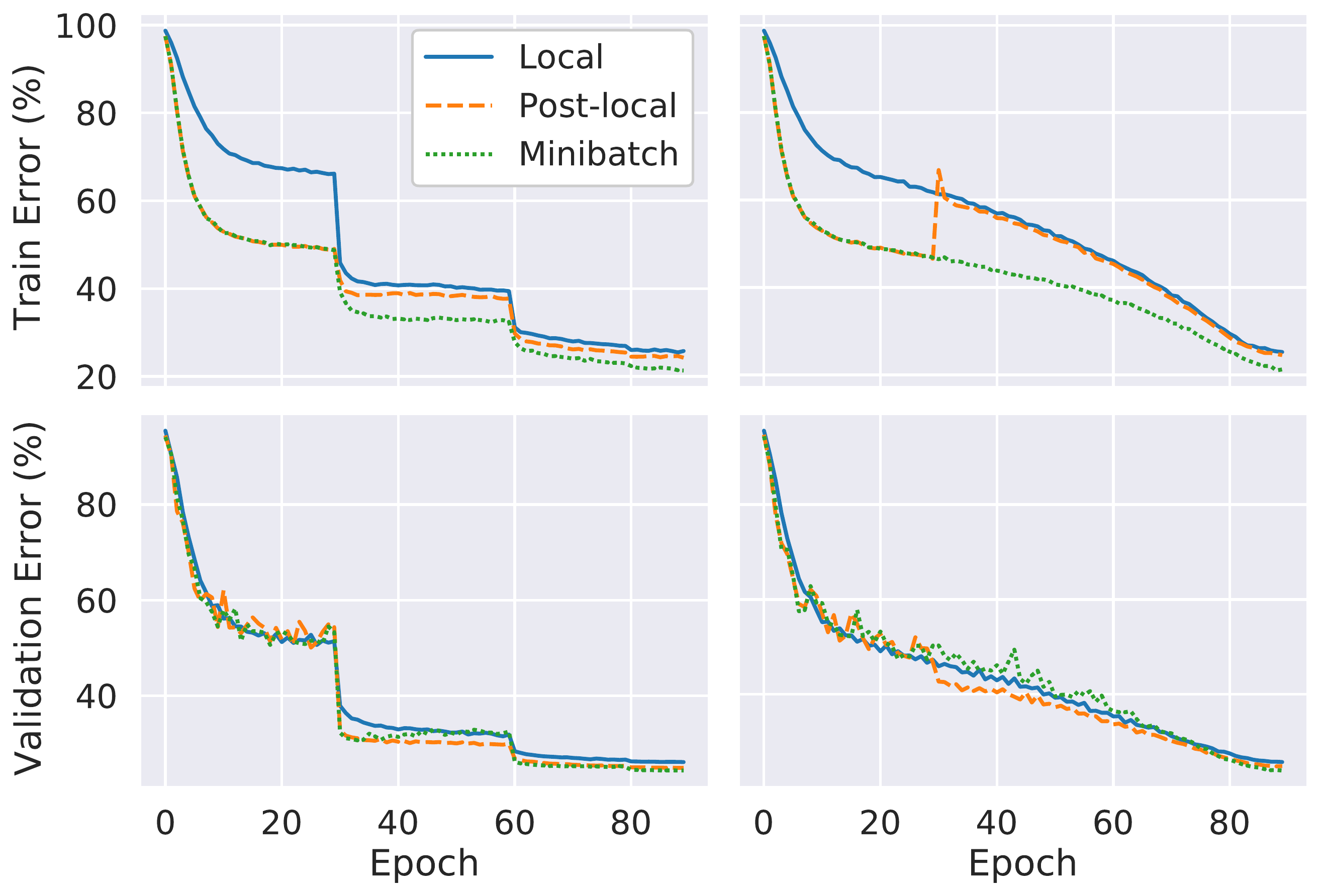}
    \caption{%
        Training and validation error minibatch, local SGD, and post-local SGD for step-wise and half cosine learning rate schedules.
        The left column is for step-wise and the right column is for half cosine.
        Results are for 64 workers and 4 local steps for local and post-local SGD.
        Switching to local SGD has a detrimental effect on the training error but a beneficial effect on the validation error.
    }
    \label{fig:T0-30_lrs_effect_per_epoch}
\end{figure}

\end{document}